\documentclass[11pt,a4paper]{article}

\usepackage[utf8]{inputenc}
\usepackage[T1]{fontenc}
\usepackage{amsmath,amssymb,amsfonts}
\usepackage{graphicx}
\usepackage{hyperref}
\usepackage{natbib}
\usepackage{booktabs}
\usepackage{xcolor}
\usepackage{authblk}
\usepackage{etoolbox}
\pretocmd{\thebibliography}{\setlength{\itemsep}{0pt}}{}{}
\makeatletter
\newcommand{\doi@aux}[1]{\appto\@currentHref{#1}}
\makeatother

\title{We Are All Creators: Generative AI, Collective Knowledge,\\and the Path Towards Human-AI Synergy}

\author[1]{Jordi Linares-Pellicer\thanks{jorlipel@upv.es}}
\author[1]{Juan Izquierdo-Domenech\thanks{juaizdom@upv.es}}
\author[1]{Isabel Ferri-Molla\thanks{isfermol@upv.es}}
\author[1]{Carlos Aliaga-Torro\thanks{calitor@upv.es}}
\affil[1]{Valencian Research Institute for Artificial Intelligence (VRAIN), Universitat Politècnica de València, Spain}

\begin{document}

\maketitle

\begin{abstract}
Generative Artificial Intelligence (GenAI) presents a profound challenge to traditional
notions of human uniqueness, particularly in the domain of creativity. Fueled by large
foundation models based on artificial neural networks (ANNs), these systems
demonstrate remarkable capabilities in generating diverse content forms, sparking
intense debate regarding authorship, copyright infringement, and the very nature of
intelligence. This paper argues that generative AI should be understood not as a
mimicry of human cognition, but as a form of alternative intelligence and alternative
creativity, operating through distinct mechanisms rooted in mathematical pattern
synthesis rather than biological understanding or verbatim replication. Examining the
analogies and crucial differences between ANNs and biological neural networks
(BNNs) reveals that AI learning is fundamentally about extracting and manipulating
statistical patterns from vast datasets, often representing a crystallized form of
collective human knowledge and expression scraped from the internet. This
perspective complicates prevailing narratives of copyright \textit{theft} and highlights the
practical and conceptual impasses in attributing AI-generated outputs to individual
sources or compensating original creators, especially given the proliferation of
open models. Rather than pursuing potentially futile regulatory or legal
restrictions, this paper advocates for a pragmatic shift towards human-AI synergy. By
embracing generative AI as a complementary tool, leveraging its alternative creative
capacities alongside human intuition, context, and ethical judgment, society can
potentially unlock unprecedented levels of innovation, democratize creative
expression, and address complex challenges across diverse fields. This collaborative
approach, grounded in a realistic understanding of AI's capabilities and limitations as
derived from collective human input, offers the most promising path forward in
navigating this technological paradigm shift. Furthermore, recognizing these models as products of collective human knowledge raises ethical considerations regarding their accessibility; ensuring equitable access to these powerful tools for knowledge transmission and learning facilitation could be crucial to prevent widening societal divides and to truly leverage their potential for collective benefit.
\end{abstract}

\textbf{Keywords:} Generative AI, Artificial Neural Networks, Creativity, Copyright, Collective
Knowledge, Human-AI Collaboration, AI Ethics, Latent Space, Open AI models

\section{Introduction}
The advent of sophisticated generative Artificial Intelligence (AI) systems, capable of producing text, images, music, video, code, etc. often indistinguishable from human creations, has ignited a global conversation about the future of creativity and the definition of authorship \citep{Stanford2025}. Foundational models, often based on complex artificial neural networks (ANNs), can generate novel output across diverse domains, seemingly encroaching upon capabilities previously considered uniquely human. This rapid technological advance has created significant social friction, manifesting primarily as intense controversies surrounding copyright law, intellectual property (IP), and the perceived threat to creative professions \citep{RAND2025}. Artists, writers, musicians, and coders grapple with the implications of AI models being trained on vast datasets scraped from the Internet, datasets often containing their copyrighted works without explicit permission or compensation \citep{arXivGenAI2025}.

This tension stems from fundamental misunderstandings about how these AI systems operate. Often treated as \textit{black boxes}, their internal mechanisms are poorly understood by the public and even by many policymakers and legal experts. This paper contends that a deeper understanding of the connectionist principles underlying generative AI, including its analogies and crucial differences with biological neural networks (BNNs), is essential for navigating the current debates constructively. We argue that generative AI does not \textit{steal} or \textit{copy}; rather, it synthesizes patterns learned from data, operating as a form of alternative intelligence and alternative creativity that shares many analogies with our own human brains. Furthermore, the data from which these models learn often represent a vast repository of collective human knowledge and expression, to which most of us have contributed.

From this perspective, the current focus on adversarial legal battles and attempts at restrictive regulation may be misplaced and ultimately impractical, particularly with the rise of powerful open models~\citep{StableDiffusion2022} that are becoming ubiquitous and easy to use. This paper proposes a shift towards a pragmatic and synergistic approach. Recognizing generative AI as a powerful, albeit different, creative partner derived from our collective digital footprint allows us to move beyond conflict toward collaboration.

This communication will explore the technical underpinnings of generative AI, contrasting its learning mechanisms with human cognition. It will explore the complexities of the copyright dilemma, examining the limitations of current legal frameworks and the impact of open models. It will develop the concept of generative AI as an embodiment of collective knowledge and creativity. Finally, it will advocate for a future focused on human-AI synergy, using these tools to enhance human capabilities, democratize creativity, and tackle complex societal challenges. This interdisciplinary exploration aligns directly with the profound cultural, social, cognitive, economic, ethical, and philosophical implications of this transformative technology from a human-centered perspective.

\section{The Connectionist Nature of Generative AI: Beyond the Black Box}
To engage meaningfully with the societal implications of generative AI, it is crucial to
move beyond simplistic metaphors and understand the fundamental principles
governing its operation. At their core, most contemporary generative models are built
upon Artificial Neural Networks (ANNs), computational structures inspired by, yet
distinct from, the biological neural networks (BNNs) found in living organisms \citep{Stanford2025}.

ANNs are essentially complex mathematical systems. The basic unit, the artificial
neuron or perceptron, is not a biological cell but a mathematical function that receives
numerical inputs, processes them (through weighted sums and activation
functions), and produces a numerical output. Instead of the electrochemical signals
used in BNNs, ANNs operate entirely on numerical representations. The "learning"
process in these networks involves adjusting a large number of parameters,
numerical values associated with the connections (weights) between artificial
neurons. These parameters effectively mirror the synaptic weights in biological neural networks,
though through distinctly mathematical rather than electrochemical mechanisms, establishing
a clear bio-inspiration despite fundamental implementation differences \citep{OpenMedScience2025}.
Through training on large datasets, these parameters, initialized
randomly, are iteratively modified (e.g., via algorithms such as backpropagation,
analogous in function to synaptic plasticity) to minimize errors and
capture underlying patterns in the data. Crucially, the resulting trained parameters embody
distributed and synergistic patterns that resemble—albeit in a purely numerical domain—the
knowledge representations and insights formed within human brains. The goal is for the network parameters to
ultimately represent a compressed, synthesized understanding of the patterns and
relationships inherent in the training corpus, encoded within a high-dimensional
"latent space" \citep{arXivLatent2025}. This perspective underscores that generative AI models
should not be conceived as mere databases or storage systems that simply retain training information,
but rather as alternative implementations of neuro-synaptic plasticity—numerical systems that
encode learned information through distributed patterns of weights, enabling novel synthesis
rather than verbatim retrieval.

Although the connectionist architecture, systems composed of interconnected
processing units, provides a compelling analogy between ANNs and BNNs \citep{Stanford2025}, the
differences are profound and critical to understanding AI's capabilities and limitations
(see Table 1). BNNs possess a staggering biological complexity, involving diverse
neuron types, intricate neurochemical signaling, developmental processes, and
constant interaction with the physical world through embodiment and sensory input
over a lifetime \citep{Frontiers2025}. ANNs, in contrast, are mathematical abstractions, typically trained on
static datasets scraped from the internet, lacking genuine embodiment, subjective
experience, or the rich, lifelong contextual learning humans acquire through sensory
interaction \citep{ResearchGate2025}. Biological learning involves multiple forms of plasticity
(experience-dependent, -independent, -expectant) and structural changes \citep{PMCNeural2025},
whereas ANN learning primarily involves optimizing numerical parameters within a
predefined architecture \citep{PMCBiological2025}.

\begin{table}[htbp]
\centering
\scriptsize
\setlength{\tabcolsep}{2pt}
\caption{Comparison of Learning Mechanisms: Biological vs. Artificial Neural Networks}
\begin{tabular}{p{1.8cm}p{5.2cm}p{5.2cm}}
\toprule
\textbf{Feature} & \textbf{Biological Neural Networks (BNNs)} & \textbf{Artificial Neural Networks (ANNs)} \\
\midrule
Basic Unit & Biological Neuron (diverse types, complex structure) & Artificial Neuron/Perceptron (Mathematical function) \\
\hline
Signal Type & Electrochemical impulses & Numerical values \\
\hline
Learning Mechanism & Synaptic Plasticity (Hebbian learning), Structural changes, Neurogenesis & Parameter/Weight Adjustment (Backpropagation, Gradient Descent) \\
\hline
Knowledge Storage & Distributed across synaptic strengths and network structure & Distributed across numerical parameters (weights, biases) \\
\hline
Information Source & Sensory input, Embodied experience, Lifelong interaction & Training Datasets (text/images from internet) \\
\hline
Context & Embodied, Situated, Biological, Evolutionary & Disembodied, Mathematical, Algorithmic \\
\hline
Underlying Principle & Biological/Electrochemical processes & Mathematical functions, Statistics \\
\bottomrule
\end{tabular}
\end{table}

This fundamental difference debunks common misconceptions fueling the copyright
debate. Generative AI models do not "store" copies of images, text passages, or
music files in the way a hard drive does. Instead, they store numerical parameters that
represent learned statistical patterns and relationships derived from the training data.
When generating output, the model uses these parameters to synthesize new data
points within the learned distribution, navigating the complex latent space that
encodes these patterns \citep{arXivLatent2025}. The process is one of synthesis and statistical inference
based on learned patterns, not retrieval of stored originals. While phenomena like
overfitting or memorization can occur, where a model does reproduce training data
verbatim (especially if data is highly duplicated or training is excessive), these are
often considered flaws or edge cases that can impair the model's ability to generalize,
rather than its primary mode of operation \citep{arXivMemorization2025}. Understanding AI as a system that learns
and synthesizes patterns via mathematical abstraction, rather than a digital
photocopier, is crucial for framing subsequent discussions about creativity and
copyright. It suggests we are dealing not with artificial human intelligence, but with a
distinct form of alternative intelligence.

\section{AI Creativity as Alternative Cognition}
If generative AI operates through pattern synthesis within a mathematical latent space
rather than human-like understanding or experience, how should we conceptualize its
creative output? Labelling it merely "artificial" risks diminishing its potential or
mischaracterizing its nature. Instead, framing it as alternative creativity stemming
from an alternative cognition offers a more productive lens.

AI creativity arises from the model's ability to navigate its learned latent space – the
high-dimensional representation of patterns extracted from training data \citep{arXivLatent2025}. By
interpolating between points in this space, combining learned features, and applying
stochastic processes, AI can generate outputs that are novel combinations of the
patterns it has absorbed. This process can yield results that appear surprising,
aesthetically pleasing, or functionally useful, meeting some criteria often associated
with creativity, such as novelty and value \citep{ResearchGateCreativity2025}. Techniques like Generative Adversarial
Networks (GANs) or diffusion models employ sophisticated methods to refine these
generated outputs, pushing the boundaries of plausible synthesis \citep{ResearchGate2025}.

However, this process differs significantly from human creativity. While it is a fallacy to
assume humans \textit{creatio ex nihilo} (out of nothing) – human creativity is deeply
influenced by experience, culture, education, and interaction with the world – human
cognition involves elements absent in current AI. These include embodied experience,
consciousness, subjective feeling, intentionality, and access to a rich context that
extends far beyond the digital data AI models are trained on \citep{ColumbiaAI2025}. Human creativity often
involves breaking established patterns based on insight or intuition derived from this
broader context, whereas AI creativity primarily operates within the patterns learned
from its data, albeit generating novel combinations thereof. While emerging techniques like reinforcement learning, open-endedness, and self-improvement aim to push models beyond the constraints of their initial training data, their exploration will remain fundamentally distinct from the rich, embodied contextual understanding that still uniquely human cognition accesses through lived experience and sensory interaction with the world.

There exists an important intersection between the capabilities of generative AI and
human creators. Both can combine existing elements in novel ways, explore variations
on themes, and generate outputs that meet certain aesthetic or functional criteria.
However, there are also significant divergences. Humans possess a depth of
understanding, ethical reasoning, and contextual awareness that AI lacks. Conversely,
AI can process and synthesize information at a scale and speed impossible for
humans, identifying patterns across vast datasets that might escape individual
perception \citep{MITSloan2025}.

Therefore, characterizing AI's output as \textit{alternative creativity} acknowledges its
distinct origins and mechanisms while recognizing its potential value. It is not a lesser
form of human creativity but a different kind, arising from mathematical optimization
and pattern manipulation rather than lived experience and biological cognition. This
framing shifts the focus from replacement to complementarity, suggesting that AI's
creative potential lies in its ability to augment and collaborate with human creativity,
rather than supplanting it \citep{Berkeley2025}.

\section{Navigating the Copyright Labyrinth: Collective Input, Individual Output?}
The capacity of generative AI to produce outputs resembling human creations, often
trained on data containing copyrighted works, lies at the heart of intense legal and
ethical conflict \citep{RAND2025}. Rights holders argue that training AI models on their works without
permission constitutes mass copyright infringement, while AI developers often invoke
defenses like fair use or specific legal exceptions \citep{arXivGenAI2025}. This complex situation is further
complicated by the nature of AI learning, the difficulty of attribution, and the rise of
open models.

The core argument from many rights holders is that AI training involves unauthorized
copying, essentially \textit{stealing} creative works to build a commercial product that may
then compete with the original creators \citep{RAND2025}. However, as discussed previously, standard
AI training does not involve storing copies of works but rather extracting patterns and encoding them as model parameters. Verbatim reproduction
(memorization) is typically an unintended byproduct, often linked to data repetition or
overfitting, and is generally avoided as it harms model generalization \citep{arXivMemorization2025}. This technical
reality challenges the straightforward \textit{copying} narrative, suggesting the
infringement analysis must be more nuanced.

In the US, the primary defense for AI developers is \textit{fair use} \citep{arXivGenAI2025}. Central to this is the
concept of \textit{transformative use} – whether the AI's use of the copyrighted material
serves a different purpose or has a different character from the original \citep{arXivGenAI2025}. Developers
argue that training AI is transformative because the goal is not to reproduce the
inputs but to learn patterns to generate entirely new outputs \citep{KluwerCopyright2025}. They might draw
parallels to cases like Google Books, where scanning books to create a searchable
database was deemed transformative \citep{KluwerCopyright2025}. However, rights holders counter that if the AI
output serves the same market as the original (e.g., AI-generated images competing
with stock photos, AI text competing with news articles), the use is substitutive, not
transformative, potentially causing market harm (the fourth fair use factor) \citep{PBWTFairUse2025}. Recent
court decisions offer mixed signals. The Supreme Court's ruling in Andy Warhol
Foundation v. Goldsmith emphasized that transformative purpose must be weighed
against commercial use and market substitution \citep{KluwerCopyright2025}. In Thomson Reuters v. Ross, a
district court rejected fair use for an AI trained on legal headnotes, finding the AI's
purpose (legal research) was too similar to the original work's purpose and potentially
harmed the market for licensing data for AI training, though this case involved
non-generative AI \citep{arXivGenAI2025}. The applicability of fair use to generative AI training remains highly
contested and fact-dependent.

The EU AI Act includes transparency requirements, such as
demanding providers publish summaries of training data, partly to help rightsholders
enforce the opt-out right \citep{OxfordTransparency2025}. However, the practical effectiveness of the opt-out and
the feasibility of creating truly informative yet non-infringing summaries remain
significant challenges \citep{arXivGenAI2025}.

The problem of attribution and compensation
presents a near-insurmountable hurdle. Tracing a specific AI output (e.g., an image, a
paragraph) back to the individual training data points that influenced its generation is
computationally complex, likely impossible in most cases \citep{OxfordAttribution2025}. Furthermore, copyright law
generally does not protect artistic \textit{style}. While AI can mimic styles, this mimicry itself may not constitute
infringement, just as human artists can be inspired by and adopt others' styles.
Lawsuits like Getty Images v. Stability AI, Andersen v. Stability AI, and Authors
Guild v. OpenAI are grappling with these issues, including claims related to the
removal or alteration of Copyright Management Information (CMI) under the DMCA
§1202 \citep{OxfordAttribution2025}.

The widespread availability of powerful open generative models, exemplified
by the release of Stable Diffusion in 2022 \citep{StableDiffusion2022}, adds another layer of complexity. These
models can be freely downloaded, modified, and fine-tuned by anyone, often using
private datasets or specific styles. Tracking the lineage of outputs generated by
potentially thousands of derivative models becomes impossible, rendering systematic
compensation or enforcement schemes based on training data provenance utterly
impractical. Users can easily train an open model (initially trained on
non-copyrighted data) with copyrighted images offline and generate new content,
making legal recourse against the original model creators or the end-users
exceedingly difficult.

This intricate legal and technical landscape suggests that seeking resolution primarily
through copyright litigation or regulation focused on individual attribution and
compensation faces formidable, perhaps insuperable, obstacles.

\section{Generative AI as Crystallized Collective Knowledge}
An alternative perspective reframes the debate by considering generative AI not merely as a technological tool but as an embodiment or crystallization of collective human knowledge and creativity.

The core idea, encapsulated in the phrase \textbf{We are all creators}—a central tenet we authors emphasize throughout this work—posits that these AI models are fundamentally built upon the vast digital output of humanity. This perspective, which we consider essential to the ongoing discourse, views generative AI systems not as independent inventors but as sophisticated processors of collective human creativity and knowledge—systems that synthesize and transform our shared digital heritage into new forms. Given this collective foundation, we argue that generative AI models should be widely accessible to prevent technological exclusion. If these systems derive their capabilities from humanity's aggregated knowledge and creativity, then restricting access to them risks creating new forms of inequality. Their potential to facilitate learning, problem-solving, and creative expression suggests an ethical imperative to ensure broad availability, particularly as these technologies become increasingly integrated into educational, professional, and creative domains.

The massive datasets used to train large foundation models are typically scraped from
the internet – encompassing websites, books, articles, images, code repositories, and
social media content \citep{OxfordAttribution2025}. This digital corpus represents an unprecedented aggregation
of human expression, knowledge, ideas, and artistic creations, contributed by billions
of individuals over decades \citep{MediumCollective2025}. When an AI model learns patterns from this data, it is, in
essence, learning from the collective intelligence and creativity embedded within it \citep{MITPhilosophy2025}.
The AI model becomes a mechanism for synthesizing, recombining, and generating
new outputs based on this immense pool of shared human input.

This \textit{collective knowledge} perspective has profound implications for the debates
surrounding ownership and compensation. If the models are fundamentally derived
from a collective, distributed input to which virtually everyone who has participated in
the digital sphere has contributed (knowingly or unknowingly), then assigning
ownership or calculating fair compensation based on individual contributions
becomes practically impossible and conceptually fraught. How could one possibly
trace the influence of billions of inputs on a single generated output (a token, a pixel)?
How would one quantify the value of each contribution – by volume, by impact, by
originality? The sheer scale and interconnectedness of the training data defy
traditional models of individual authorship and reward.

The rise of open AI models further reinforces this collective dimension \citep{StableDiffusion2022}.
Models like Stable Diffusion, once released, become part of a shared technological
commons, accessible for anyone to use, study, modify, and build upon. This
accelerates innovation but also diffuses responsibility and control, making centralized
compensation or restriction schemes even less feasible. The continuous fine-tuning
and adaptation of these models by a global community further interlace individual
and collective contributions.

Viewing generative AI through this lens does not negate the validity of copyright or
the importance of creator rights, but it suggests that the existing frameworks may be
ill-suited to this new reality. It shifts the focus from individual \textit{theft} to the complex
dynamics of collective creation and technological mediation. It also brings other
ethical dimensions to the forefront, moving beyond copyright.

\section{Towards Human-AI Synergy: A Pragmatic Path Forward}
Given the technical nature of generative AI, the complexities of the legal landscape, the impracticality of individual attribution for collective inputs, and the unstoppable momentum driven by user demand and open availability, a purely restrictive or litigious approach seems destined to fail. A more pragmatic and potentially fruitful path lies in embracing human-AI synergy, viewing these technologies not as adversaries but as powerful, complementary tools capable of augmenting human capabilities. Global access to these models promises substantial societal benefits: democratization of creative and analytical capabilities across socioeconomic boundaries; acceleration of scientific discovery and innovation through widespread computational assistance; enhancement of educational opportunities in resource-limited settings; preservation and revitalization of cultural knowledge through accessible digital interfaces; and fostering cross-cultural understanding through reduced language barriers. Indeed, this global accessibility seems ethically warranted given that these models are fundamentally trained on and learn from humanity's collective intellectual and creative output. If the source material represents our shared cultural heritage and collective knowledge, then shouldn't the technologies derived from this collective resource be similarly accessible to all? By ensuring inclusive access rather than concentrated control, we can potentially harness these technologies to address collective challenges while minimizing the risk of exacerbating existing inequalities in technological access and literacy.

The reality is that generative AI models, both proprietary and open, are widely
available and integrated into numerous workflows and creative processes. Users,
from individuals exploring creative outlets to professionals seeking productivity gains,
desire these tools and will likely continue to use them, turning to open
alternatives if proprietary systems become overly restricted or expensive. Attempts to
legislate strict controls or enforce universal compensation schemes face enormous
practical challenges, including the difficulty of monitoring decentralized open
usage and the sheer complexity of global copyright law \citep{arXivGenAI2025}.

Therefore, a pragmatic approach involves shifting the focus from prohibition and
conflict to collaboration and integration. This means recognizing AI's \textit{alternative}
intelligence and creativity as distinct from, but potentially synergistic with, human
cognition \citep{MITSloan2025}. AI excels at processing vast data, identifying patterns, generating
variations rapidly, and automating repetitive tasks \citep{MITSloan2025}. Humans excel at contextual
understanding, ethical judgment, nuanced communication, emotional intelligence, and
truly novel conceptual leaps \citep{MITSloan2025}.

The synergy arises when these complementary strengths are combined \citep{SmythOS2025}. In creative
fields, AI can act as a tireless brainstorming partner, a generator of initial drafts, a
tool for exploring stylistic variations, or a means of automating laborious aspects of
production, freeing human creators to focus on higher-level ideation, refinement, and
emotional expression. This collaboration has the potential to democratize creativity,
empowering individuals who lack traditional skills or resources to bring their ideas to
life.

Beyond the arts, human-AI synergy holds immense promise for science, technology,
and complex problem-solving \citep{MITSloan2025}. AI can analyze massive datasets in fields like medicine
or climate science far faster than humans, identifying potential drug candidates,
predicting disease outbreaks, or modeling complex environmental systems. Human
researchers can then interpret these findings, design experiments, apply ethical
considerations, and guide the research direction \citep{CambridgeCollective2025}. This collaborative approach can
accelerate discovery and lead to solutions for pressing global challenges.

This pragmatic vision does not ignore the legitimate concerns surrounding AI. Issues
of job displacement, bias amplification, misinformation, privacy, and the
concentration of power require ongoing attention and the development of robust
ethical frameworks and governance structures \citep{MITPhilosophy2025}. However, these frameworks should
aim to guide the development and deployment of AI towards beneficial collaboration,
rather than attempting to halt its progress or become entangled in intractable IP
disputes based on potentially outdated paradigms. The goal should be to harness AI's
potential while mitigating its risks, fostering a future where human and alternative
intelligence work together.

\section{Conclusion}
Generative AI represents a significant technological inflection point, challenging our
conceptions of creativity, intelligence, and authorship. Understanding
generative AI as a form of \textit{alternative intelligence and creativity}, one that excels at synthesizing
complex patterns learned from vast datasets derived from collective human
knowledge with important analogies with our own brains, provides a more accurate foundation for navigating its societal
implications.

This perspective reveals the profound difficulties in applying traditional copyright
frameworks, designed for individual human creators, to outputs generated from
diffuse, collective inputs via mathematical abstraction. The practical and conceptual
barriers to tracing attribution and implementing fair compensation, particularly in the
face of powerful and freely adaptable open models, suggest that legal battles
centered on infringement for training data may prove ultimately intractable and
perhaps counterproductive.

A more pragmatic and forward-looking approach, as advocated in this paper, involves
embracing the inevitability of these tools and focusing on harnessing their potential
through human-AI synergy. By recognizing the complementary strengths of human
intuition, context, and ethical judgment alongside AI's capacity for scale, speed, and
pattern manipulation, we can foster collaborations that augment human capabilities,
democratize creative expression, and accelerate progress in science and technology.
This requires moving beyond an adversarial mindset towards one of integration and
co-evolution.

This paradigm shift is not without challenges. Concerns regarding bias, labor
displacement, misinformation, and ethical governance are real and demand careful
consideration and proactive solutions. However, attempting to stifle the technology or
force it into ill-fitting legal boxes is unlikely to succeed and risks forfeiting its
considerable potential benefits. The path forward lies in developing robust ethical
guidelines, fostering AI literacy, and cultivating collaborative practices that leverage
this powerful new form of alternative intelligence – born from our collective past – to
build a more creative, productive, and equitable future. This requires a societal dialogue rooted in a clear understanding of what generative AI is, and is not. Crucially, moving forward wisely necessitates addressing the accessibility of this new paradigm. If generative AI is indeed a crystallization of collective human knowledge, there is a strong ethical argument for ensuring broad and equitable access to its benefits, particularly its capacity to democratize creativity and facilitate learning. Preventing the concentration of this power and bridging potential digital divides should be a key component of any strategy aiming for genuine human-AI synergy and societal advancement. There is no turning back; the challenge lies in moving forward wisely.

\bibliographystyle{plainnat}
\bibliography{bibliography}

\end{document}